\pdfoutput=1
\documentclass[lettersize,journal]{IEEEtran}
\usepackage{amsmath,amsfonts}
\usepackage{array}
\usepackage[caption=false,font=footnotesize,labelfont=normalfont,textfont=normalfont]{subfig}
\usepackage{textcomp}
\usepackage{stfloats}
\usepackage{url}
\usepackage{verbatim}
\usepackage{graphicx}
\usepackage{cite}
\usepackage{booktabs}
\newsavebox{\panelbox}
\usepackage{multirow}
\usepackage{amssymb}
\usepackage{xcolor}

\begin{document}

\title{GS-Net: Heterogeneous Vehicle Data Reuse via Generalizable Plug-and-Play 3DGS Module}

\author{Yichen Zhang$^{1,2,3,\dagger}$, Zihan Wang$^{1,2,\dagger}$, Jiali Han$^{4}$, Peilin Li$^{4}$, Jiaxun Zhang$^{1,2,5}$, Jianqiang Wang$^{1,2}$\\ Lei He$^{1,2,*}$ and Keqiang Li$^{1,2}$% <-this % stops a space
\thanks{This work is supported by National Key R\&D Program of China 2024YFB2505500.}% <-this % stops a space
\thanks{$^{\dagger}$These authors contributed equally to this work.}%
\thanks{$^{1}$The School of Vehicle and Mobility, Tsinghua University, China}%
\thanks{$^{2}$The State Key Laboratory of Intelligent Green Vehicle and Mobility, Tsinghua University, China}%
\thanks{$^{3}$The College of AI, Tsinghua University, China}%
\thanks{$^{4}$Independent Researcher}%
\thanks{$^{5}$University of Illinois at Urbana-Champaign, USA}%
\thanks{*Corresponding author: Lei He (helei2023@tsinghua.edu.cn).}%
}

\maketitle

\begin{abstract}
 
\textcolor{black}{End-to-end autonomous driving is increasingly data-driven, yet data reuse across vehicles remains limited. Each new vehicle often requires additional data collection and retraining because camera translation, orientation, and field of view differ across sensor layouts. Cross-sensor view synthesis offers a promising route for cross-platform data reuse by synthesizing images under novel sensor configurations from existing sensor data. To realize this goal, we propose GS-Net, a lightweight plug-and-play module that aggregates local geometric context from sparse Structure-from-Motion (SfM) point clouds and expands each point into multiple dense Gaussian primitives in a single forward pass, learning a cross-scene generalizable initialization for standard 3DGS that improves rendering quality for both interpolated views along the original sensor trajectories and extrapolated views at new sensor positions. In such settings, target camera viewpoints often lie beyond the convex hull of training cameras, whereas existing methods and benchmarks predominantly evaluate in-hull interpolation, leaving the extrapolation regime underexplored. To enable quantitative evaluation under this setting, we introduce CARLA-NVS, the first benchmark explicitly designed for cross-sensor view synthesis. Unlike existing autonomous driving datasets that typically employ no more than eight cameras in fixed configurations, CARLA-NVS features 12 cameras uniformly distributed at $30^{\circ}$ azimuth intervals, enabling controlled evaluation of both interpolated and extrapolated viewpoints. Experiments on CARLA-NVS show that GS-Net improves rendering quality by 2.08\,dB PSNR on interpolated views and 1.86\,dB on extrapolated views over standard 3DGS, while achieving a $50\times$ faster initialization compared to MVS-based densification. These results offer a practical step toward scalable cross-vehicle data reuse in autonomous driving.}
 
\end{abstract}


\section{Introduction}

\IEEEPARstart{D}{ata-driven} autonomous driving relies on large-scale datasets, whose collection and annotation are expensive and time-consuming. In practice, most driving datasets are collected with a fixed sensor layout, inherently tying the captured data to that configuration. When a new vehicle platform adopts a different sensor layout, the previously collected data cannot be straightforwardly reused, forcing a costly cycle of re-collection and re-annotation. This poor data reusability across sensor layouts has become a key bottleneck for scaling autonomous driving systems across heterogeneous vehicle platforms.

To enable scalable data reuse, cross-sensor view synthesis is a promising paradigm that aims to render images for a target sensor layout using recordings collected under a different layout. Unlike conventional novel view synthesis, where target viewpoints typically lie within the convex hull of the training cameras, cross-sensor transfer fundamentally alters camera extrinsics. Consequently, target viewpoints frequently lie outside this hull, making pose extrapolation a core requirement. To achieve such transfer in large-scale outdoor environments, recent advances in 3D Gaussian Splatting (3DGS)~\cite{c2} offer a powerful renderable scene representation. With its efficient and photorealistic rendering, 3DGS provides a practical foundation for this task.

\textcolor{black}{However, applying 3DGS to cross-sensor synthesis exposes a practical limitation of its standard initialization pipeline. SfM~\cite{schoenberger2016sfm} point clouds from driving sequences are inherently sparse, and the subsequent heuristic Gaussian initialization offers limited geometric coverage. When target cameras are placed at extrapolated positions, they observe regions poorly covered by the training cameras, and per-scene optimization is weakly constrained and often fails to recover faithful geometry in these areas. Meanwhile, the underlying local mapping from sparse SfM points to dense Gaussian primitives exhibits strong structural consistency, enabling it to be learned and transferred across diverse driving scenes. To exploit this transferable geometric prior, we propose GS-Net, a lightweight module that predicts dense, high-quality Gaussian primitives directly from sparse SfM inputs in a single forward pass. The predicted initialization is then refined by the standard 3DGS optimization process, improving rendering quality for both interpolated views along the original sensor trajectories and extrapolated views at new sensor positions.}

\textcolor{black}{To evaluate this setting rigorously, cross-sensor synthesis also requires ground-truth images at alternative camera positions. Existing autonomous driving datasets, such as KITTI~\cite{geiger2012kitti}, nuScenes~\cite{caesar2020nuscenes}, and Waymo~\cite{sun2020waymo}, provide strictly fixed sensor setups and therefore lack observations at new camera placements. Related benchmarks like EUVS~\cite{EUVS} study extrapolation across driving trajectories, but they vary vehicle poses rather than sensor configurations. To enable quantitative evaluation of cross-sensor view synthesis, we introduce CARLA-NVS, a controlled benchmark designed to provide multi-view ground truth under diverse sensor layouts. It supports systematic evaluation of both interpolation and extrapolation under sensor-layout changes.}

To address these challenges, we make the following contributions:
\begin{itemize}
    \item[1)] \textcolor{black}{We propose GS-Net, a lightweight plug-and-play module that learns cross-scene geometric priors to predict dense Gaussian primitives from sparse SfM inputs in a single forward pass, achieving high-quality initialization for standard 3DGS.}
    \item[2)] \textcolor{black}{We introduce CARLA-NVS, the first benchmark explicitly designed for cross-sensor view synthesis in autonomous driving. It features 12 cameras at $30^{\circ}$ azimuth intervals for controlled evaluation of both interpolation and extrapolation under sensor layout changes, and is released together with a scalable data engine and all processed geometric representations.}
    \item[3)] \textcolor{black}{Extensive experiments on CARLA-NVS demonstrate that GS-Net improves rendering quality under both interpolated and extrapolated viewpoints, while downstream perception consistency further supports its utility for cross-vehicle data reuse.}
\end{itemize}

\section{Related Work}

\subsection{View Synthesis Evaluation in Autonomous Driving}
Novel view synthesis in autonomous driving has advanced rapidly with methods ranging from NeRF~\cite{c13} to 3DGS~\cite{c2}, yet evaluation protocols have not kept pace. The most common paradigm is temporal interpolation, where training and test images come from the same cameras and differ only in timestamp. The EUVS benchmark~\cite{EUVS} goes a step further by evaluating on views from entirely different driving trajectories, introducing spatial extrapolation into the evaluation. However, neither paradigm addresses the cross-sensor scenario, where the driving trajectory remains the same but the camera configuration differs across vehicle platforms. A fundamental reason is that existing datasets do not provide ground-truth images from alternative camera placements. KITTI~\cite{geiger2012kitti}, nuScenes~\cite{caesar2020nuscenes}, and Waymo~\cite{sun2020waymo} each equip a vehicle with a single fixed sensor configuration, making cross-sensor evaluation infeasible. To overcome this limitation, the proposed CARLA-NVS dataset augments a conventional six-camera surround-view layout with six additional cameras at $30^{\circ}$ angular offsets, forming a 12-camera configuration that naturally produces ground-truth images from alternative camera placements for cross-sensor evaluation.
 
\subsection{\textcolor{black}{Feed-forward Gaussian Splatting}}
\textcolor{black}{Feed-forward Gaussian Splatting methods learn to predict 3D Gaussians directly from input images in a single forward pass. For instance, pixelSplat~\cite{pixelsplat} regresses pixel-aligned Gaussians from an image pair, sampling their centers from a predicted depth distribution, and MVSplat~\cite{mvsplat} from a multi-view cost volume. Later works add monocular-depth priors~\cite{monosplat}, pose-free training~\cite{ggrt}, richer feature or semantic cues~\cite{sparsplat,stereogs,c3gs,gssplat}, and generalizable surface reconstruction~\cite{meshsplat}. However, predicting the Gaussians in one pass leaves no per-scene refinement, so their fidelity is bounded, and the predicted geometry grows unreliable far from the input views. A single feed-forward pass is therefore insufficient under large viewpoint extrapolation.}

\subsection{\textcolor{black}{Gaussian Primitive Initialization}}
\textcolor{black}{The rendering quality of 3DGS is closely tied to its Gaussian initialization, which the standard pipeline derives from sparse SfM points before per-scene optimization with adaptive density control. In textureless or weakly observed regions where SfM is sparse, the initialization is unreliable and leaves artifacts that optimization alone cannot resolve. Some methods inject geometric cues into the point cloud, such as planar propagation~\cite{cheng2024Gaussianpro} and epipolar depth~\cite{nexusgs}, while others learn initialization or optimization from data~\cite{quicksplat,instantsplat,g3r,fastgs}. The former usually rely on per-scene heuristics rather than a learned cross-scene prior. Among learned Gaussian initialization methods, QuickSplat~\cite{quicksplat} is closely related, but it mainly targets indoor surface reconstruction and relies on accurate geometric supervision from high-fidelity scanned mesh geometry, which is generally unavailable for large-scale outdoor driving scenes. In contrast, GS-Net learns a cross-scene Gaussian initialization prior from image-derived pseudo-ground-truth Gaussians, without requiring ground-truth meshes, and leaves the standard 3DGS per-scene optimization unchanged. Trained across driving scenes, it generalizes to unseen environments and can be used as a plug-and-play initialization module for existing 3DGS pipelines.}

\begin{figure*}[t]
    \centering
    \includegraphics[width=0.8\linewidth]{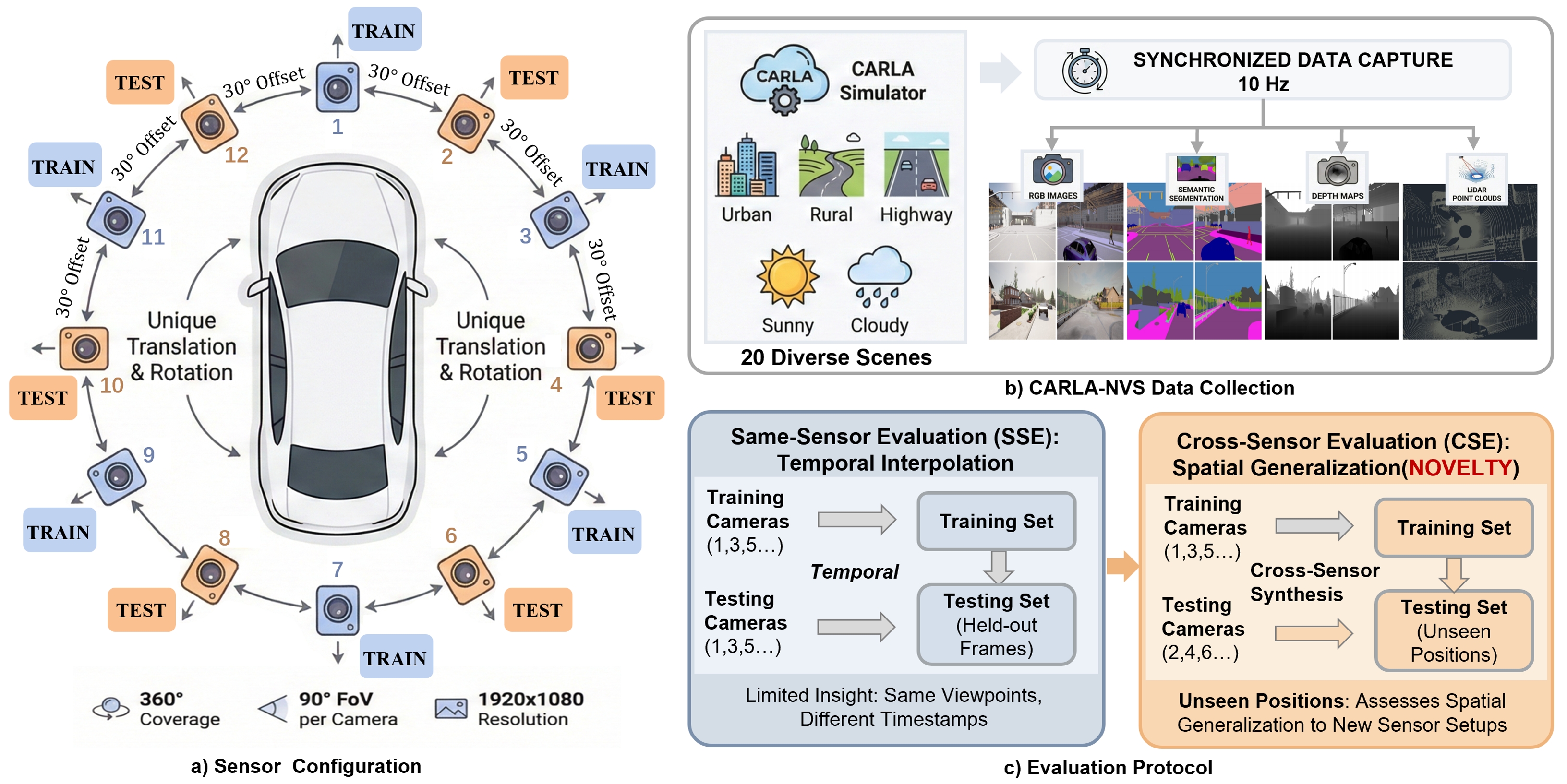}
    \caption{Overview of the CARLA-NVS dataset. (a) Sensor configuration with 12 cameras around the vehicle, where odd-numbered cameras form the training set and even-numbered the testing set, each pair offset by $30^{\circ}$. (b) Data collection over 20 scenes across urban, rural, and highway environments under varying weather, with synchronized RGB, semantic segmentation, depth, and LiDAR at 10\,Hz. (c) Evaluation protocols, where Same-Sensor Evaluation tests temporal interpolation on held-out frames from training cameras and Cross-Sensor Evaluation tests spatial generalization to unseen camera positions.}
    \label{fig:carla_nvs}
\end{figure*}

\section{Preliminaries}
\label{sec:preliminaries}

3DGS~\cite{c2} models the 3D scene as a set of anisotropic 3D Gaussians, which are rendered to images using a splatting-based rasterization technique. For each Gaussian primitive $g_n$, its spatial influence is defined by the Gaussian function:
\begin{equation}
G_n(\mathbf{x}) = \exp\!\left(-\frac{1}{2}(\mathbf{x} - \boldsymbol{\mu}_n)^\top \boldsymbol{\Sigma}_n^{-1} (\mathbf{x} - \boldsymbol{\mu}_n)\right),
\end{equation}
where $\boldsymbol{\mu}_n \in \mathbb{R}^{3}$ refers to its mean position and $\boldsymbol{\Sigma}_n \in \mathbb{R}^{3\times3}$ refers to its covariance matrix. To ensure the positive semi-definite property of the covariance matrix during optimization, it is further expressed as $\boldsymbol{\Sigma}_n = \mathbf{R}_n \mathbf{S}_n \mathbf{S}_n^\top \mathbf{R}_n^\top$, where the rotation matrix $\mathbf{R}_n \in \mathbb{R}^{3\times3}$ is orthogonal and the scale matrix $\mathbf{S}_n \in \mathbb{R}^{3\times3}$ is diagonal. Each Gaussian additionally carries a learned opacity $\alpha_n$ and a color $\mathbf{C}_{\text{rgb}_n}$ represented via spherical harmonics.

To render an image from a given viewpoint, the color of each pixel $p$ is calculated by blending $N$ ordered Gaussians $\{g_n \mid n = 1, \dots, N\}$ overlapping $p$ as:
\begin{equation}
\mathbf{C}(p) = \sum_{n=1}^{N} \mathbf{C}_{\text{rgb}_n}\, \alpha_n \prod_{j=1}^{n-1} (1 - \alpha_j),
\end{equation}
where Gaussians that cover $p$ are sorted in ascending order of their depths under the current viewpoint. Through differentiable rendering, all attributes of the Gaussians can be optimized end-to-end via training view reconstruction. The standard 3DGS framework initializes these primitives from sparse SfM point clouds using heuristic parameters and refines them through per-scene photometric optimization.

\section{The CARLA-NVS Dataset}
 
To provide the first dedicated benchmark for cross-sensor view synthesis, we construct the CARLA-NVS dataset using the CARLA simulator~\cite{dosovitskiy2017carla}.
 
\subsection{Data Collection}

CARLA-NVS is collected with a 12-camera vehicle rig in the CARLA simulator, as shown in Fig.~\ref{fig:carla_nvs}(a). The cameras are mounted at distinct positions around the vehicle perimeter and oriented at $30^{\circ}$ azimuth intervals, providing full $360^{\circ}$ coverage with a $90^{\circ}$ field of view at a resolution of $1920\times1080$. In addition to RGB images, the simulator records pixel-aligned depth maps, semantic segmentation masks, and point clouds from a 128-channel LiDAR. Using this sensor configuration, we collect 20 diverse driving scenes covering urban streets, rural roads, and highways under both sunny and cloudy conditions, as illustrated in Fig.~\ref{fig:carla_nvs}(b). For each scene, all 12 cameras record synchronously at 10\,Hz over a 10-second driving segment covering approximately 100 meters. We divide each scene into 10 one-second sequences, each containing 120 RGB images from all cameras. In total, CARLA-NVS contains 24{,}000 RGB images with aligned depth and semantic annotations, together with dense LiDAR point clouds. RGB and annotation maps are stored in PNG format, and point clouds are stored in PCD format.

% Place this figure at the appropriate location in your paper.
% Since it spans the full page width, use figure* environment.

\begin{figure*}[t]
    \centering
    \includegraphics[width=0.8\linewidth]{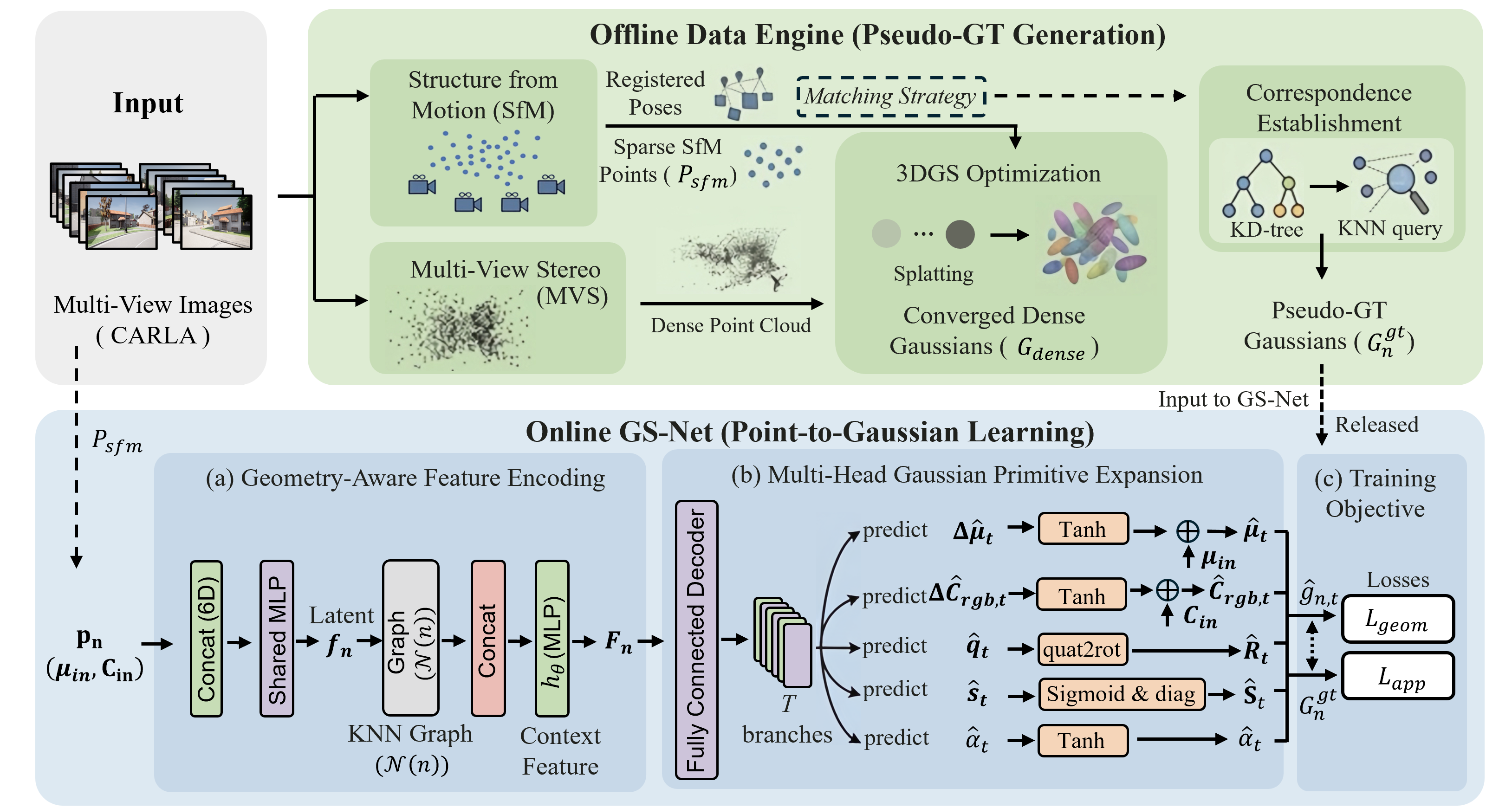}
    \caption{Overview of the proposed framework. \textbf{Top}: The offline data engine constructs pseudo-ground-truth targets by first extracting sparse SfM point clouds $\mathcal{P}_{\text{sfm}}$ and dense MVS reconstructions, then optimizing dense Gaussian primitives $\mathcal{G}_{\text{dense}}$ via 3DGS, and finally establishing sparse-to-dense correspondences $\mathcal{G}^{gt}_n$ through KD-tree nearest-neighbor queries. \textbf{Bottom}: The online GS-Net takes $\mathcal{P}_{\text{sfm}}$ as input and performs point-to-Gaussian learning through three stages: (a) Geometry-Aware Feature Encoding aggregates local neighborhood context into representations $\mathbf{F}_n$; (b) Multi-Head Gaussian Primitive Expansion predicts $T$ regularized Gaussian primitives per input point via incremental offsets and tailored activations; (c) Training Objective supervises the predicted primitives $\hat{g}_{n,t}$ against the pseudo-ground-truth $\mathcal{G}^{gt}_n$ using decoupled geometric and appearance losses.}
    \label{fig:framework}
\end{figure*}

\subsection{Data Engine}
\label{sec:data_engine}
 
To support future research on novel view synthesis, CARLA-NVS goes beyond distributing raw sensor captures. We construct a scalable, automated offline data engine, as illustrated in the top part of Fig.~\ref{fig:framework}, to systematically generate standardized geometric representations and dense point-to-Gaussian correspondences. Specifically, the engine pipeline consists of three sequential stages. First, COLMAP~\cite{schoenberger2016sfm} is used to perform Structure-from-Motion (SfM), producing sparse point clouds $\mathcal{P}_{\text{sfm}}$ and registered camera poses. Second, Multi-View Stereo (MVS) algorithms~\cite{furukawa2009mvs,schoenberger2016mvs} generate dense point cloud reconstructions from the registered images. Third, these dense reconstructions serve as the initialization for per-scene 3D Gaussian Splatting (3DGS) optimization~\cite{c2}, where each sequence undergoes 30{,}000 iterations to produce converged, high-quality dense Gaussian primitives, denoted as $\mathcal{G}_{\text{dense}}$. To further establish dense correspondences between the sparse SfM inputs and the optimized Gaussian representations, a KD-tree is constructed over the center positions of $\mathcal{G}_{\text{dense}}$. For each sparse SfM point in $\mathcal{P}_{\text{sfm}}$, its $K$ nearest dense Gaussian primitives are retrieved as the corresponding targets.
\subsection{Evaluation Protocol}
\label{sec:eval_protocol}
 
The 12 cameras in CARLA-NVS are designed to simulate cross-platform sensor layout differences. The 6 odd-numbered cameras \{1, 3, 5, 7, 9, 11\}, uniformly spaced at $60^{\circ}$ intervals, represent a typical surround-view configuration of a source vehicle, while the 6 even-numbered cameras \{2, 4, 6, 8, 10, 12\} represent an alternative target configuration offset by $30^{\circ}$. Based on this design, we define two evaluation settings, as shown in Fig.~\ref{fig:carla_nvs}(c). Same-Sensor Evaluation (SSE) assesses rendering quality within the source sensor layout, where only the 6 source cameras are involved. Within each sequence, 8 frames per camera are used for scene optimization and the remaining 2 are held out for testing, yielding 48 optimization images and 12 test images per sequence. This setting evaluates temporal interpolation along the observed camera trajectories. Cross-Sensor Evaluation (CSE) directly targets the cross-sensor data reuse scenario. Within each sequence, all frames from the 6 source cameras are used for scene reconstruction, totaling 60 images, while the 60 images from the 6 target cameras are reserved for evaluation. This setting assesses whether a method can synthesize high-quality images at camera positions entirely absent during reconstruction.

\section{METHOD}

In this paper, we propose GS-Net, a lightweight and plug-and-play module that predicts dense Gaussian primitives directly from sparse Structure-from-Motion (SfM) point clouds for cross-sensor view synthesis. The pipeline of our GS-Net is illustrated in the bottom part of Fig.~\ref{fig:framework}. Specifically, given sparse SfM points as input, GS-Net first extracts local geometric priors through a geometry-aware encoder (Sec.~\ref{sec:encoding}) that aggregates neighborhood information into context-aware feature representations. The multi-head expansion stage (Sec.~\ref{sec:expansion}) subsequently maps these features to dense Gaussian primitives, employing tailored activation functions and an incremental prediction strategy to guarantee physically valid outputs. Finally, the training objective (Sec.~\ref{sec:training}) leverages the dense correspondences generated by the offline data engine (Sec.~\ref{sec:data_engine}) as pseudo-ground-truth supervision.

\subsection{Geometry-Aware Feature Encoding}
\label{sec:encoding}

The geometry-aware encoder aggregates the local geometric context around each sparse input point, as shown in Fig.~\ref{fig:framework}(a). Specifically, given a set of sparse SfM points $\mathcal{P}_{\text{sfm}} = \{\mathbf{p}_n\}_{n=1}^{N}$, each point $\mathbf{p}_n$ carries a position $\boldsymbol{\mu}_{\text{in}} \in \mathbb{R}^3$ and a color $\mathbf{C}_{\text{in}} \in \mathbb{R}^3$. For each point, these two attributes are concatenated into a 6-dimensional raw feature and embedded into a latent representation $\mathbf{f}_n \in \mathbb{R}^{d}$ through a weight-shared multilayer perceptron.

Subsequently, we construct an $M$-nearest neighbor graph $\mathcal{N}(n) = \{j_1, \dots, j_M\}$ for each point based on Euclidean distance. To integrate local geometric context, the feature of the center point is concatenated with the features of its $M$ neighbors and mapped to a context-aware representation. This process can be formulated as:
\begin{equation}
\mathbf{F}_n = h_\Theta\!\left( \big[ \mathbf{f}_n;\, \mathbf{f}_{j_1};\, \dots;\, \mathbf{f}_{j_M} \big] \right),
\end{equation}
where $[\,\cdot\,;\,\cdot\,]$ denotes concatenation, $h_\Theta$ denotes a multilayer perceptron with learnable parameters $\Theta$, and $\mathbf{F}_n \in \mathbb{R}^{D}$ is the context-aware representation. Aggregating neighborhood context enables the network to implicitly capture local surface structures, providing adequate contextual information for predicting geometrically consistent Gaussian primitives in the subsequent expansion stage.

\subsection{Multi-Head Gaussian Primitive Expansion}
\label{sec:expansion}

The expansion stage is central to our framework, as it transforms each sparse input point into $T$ dense Gaussian primitives, where $T$ is a tunable expansion factor, to provide sufficient geometric support for rendering.

As shown in Fig.~\ref{fig:framework}(b), for each input point $\mathbf{p}_n$, the context-aware representation $\mathbf{F}_n$ is fed into a fully connected decoder that predicts the complete set of Gaussian attributes defined in Sec.~\ref{sec:preliminaries} for each of its $T$ expanded primitives. Since directly regressing position and color in absolute coordinates can lead to spatial drift and training instability, we adopt an incremental prediction strategy that formulates them as bounded residual offsets added to the input attributes of $\mathbf{p}_n$. Meanwhile, the covariance matrix is constructed from the predicted rotation quaternion and scale vector through tailored activation functions, following the factorization in Sec.~\ref{sec:preliminaries}. Formally, for the $t$-th expanded primitive where $t \in \{1, \dots, T\}$, we denote all network-predicted quantities with $\hat{\cdot}$, and use $\Delta$ to indicate offsets relative to the input attributes of $\mathbf{p}_n$. The regularized attributes are computed as:
\begin{equation}
\begin{aligned}
\hat{\boldsymbol{\mu}}_{t} &= \boldsymbol{\mu}_{\text{in}} + \mathrm{Tanh}(\Delta\hat{\boldsymbol{\mu}}_t), \\
\hat{\mathbf{C}}_{\text{rgb},t} &= \mathbf{C}_{\text{in}} + \mathrm{Tanh}(\Delta\hat{\mathbf{C}}_{\text{rgb},t}), \\
\hat{\mathbf{R}}_t &= \mathrm{quat2rot}\!\left(\hat{\mathbf{q}}_t \,/\, \lVert \hat{\mathbf{q}}_t \rVert_2 \right), \\
\hat{\mathbf{S}}_t &= \mathrm{diag}\!\left(\sigma(\hat{\mathbf{s}}_t)\right), \\
\hat{\boldsymbol{\Sigma}}_{t} &= \hat{\mathbf{R}}_t \, \hat{\mathbf{S}}_t \, \hat{\mathbf{S}}_t^\top \hat{\mathbf{R}}_t^\top,
\end{aligned}
\end{equation}
where $\Delta\hat{\boldsymbol{\mu}}_t$ and $\Delta\hat{\mathbf{C}}_{\text{rgb},t}$ are the predicted position and color offsets, $\hat{\mathbf{q}}_t$ is the predicted rotation quaternion, $\hat{\mathbf{s}}_t$ is the predicted scale vector, $\sigma(\cdot)$ denotes the Sigmoid function, and $\mathrm{quat2rot}(\cdot)$ converts a unit quaternion to a rotation matrix. Through these activation functions, position offsets are bounded to $(-1, 1)$ by Tanh to firmly anchor the densified primitives near the sparse input, while Sigmoid constrains scale entries to strictly positive values to prevent degenerate zero-volume primitives. The opacity $\hat{\alpha}_t$ is independently regulated by a Tanh activation mapping to $(-1, 1)$, where primitives with negative opacity are naturally discarded during rendering. Note that $\Delta\hat{\mathbf{C}}_{\text{rgb},t}$ corresponds to the zero-order spherical harmonics coefficient for view-independent diffuse color, while higher-order coefficients are left to the subsequent per-scene optimization.

\subsection{Training Objective}
\label{sec:training}

With the predicted primitives from the expansion stage and the paired targets provided by the offline data engine (Sec.~\ref{sec:data_engine}), we formulate the training objective as illustrated in Fig.~\ref{fig:framework}(c). For each sparse SfM point $\mathbf{p}_n$, we denote its $K$ nearest dense Gaussian primitives as the pseudo-ground-truth set $\mathcal{G}^{gt}_n = \{g^{gt}_{n,1}, \dots, g^{gt}_{n,K}\}$, where each $g^{gt}_{n,t}$ is a complete Gaussian primitive with attributes defined in Sec.~\ref{sec:preliminaries}, and the superscript $gt$ denotes ground-truth values throughout this section. By setting the network expansion factor to $T = K$, the $t$-th predicted primitive $\hat{g}_{n,t}$ of point $\mathbf{p}_n$ is directly paired with its target $g^{gt}_{n,t}$ for supervision.

To explicitly decouple the physical semantics of the Gaussian attributes, the loss for each predicted-target pair is divided into a geometric term and an appearance term. The geometric loss supervises the position offset, rotation matrix, and scale matrix:
\begin{equation}
\mathcal{L}_{\text{geom}} = \lVert \Delta\hat{\boldsymbol{\mu}}_{t} - \Delta\boldsymbol{\mu}^{gt}_{t} \rVert_2^2 + \lVert \hat{\mathbf{R}}_{t} - \mathbf{R}^{gt}_{t} \rVert_F^2 + \lVert \hat{\mathbf{S}}_{t} - \mathbf{S}^{gt}_{t} \rVert_F^2,
\end{equation}
where the ground-truth position offset is computed as $\Delta\boldsymbol{\mu}^{gt}_{t} = \boldsymbol{\mu}^{gt}_{t} - \boldsymbol{\mu}_{\text{in}}$ to align with the incremental prediction design. The appearance loss supervises the color offset and opacity:
\begin{equation}
\mathcal{L}_{\text{app}} = \lVert \Delta\hat{\mathbf{C}}_{\text{rgb},t} - \Delta\mathbf{C}^{gt}_{\text{rgb},t}\rVert_2^2 + \left(\max\!\left(\hat{\alpha}_{t},\, 0\right) - \alpha^{gt}_{t}\right)^2,
\end{equation}
where $\Delta\mathbf{C}^{gt}_{\text{rgb},t} = \mathbf{C}^{gt}_{\text{rgb},t} - \mathbf{C}_{\text{in}}$, and the $\max(\hat{\alpha}_{t}, 0)$ operation clips the Tanh-activated opacity, ensuring that only primitives with positive opacity contribute to the supervision signal. The total loss of GS-Net is obtained by averaging over all $N$ input points and their $T$ expanded primitives:
\begin{equation}
\mathcal{L} = \frac{1}{NT} \sum_{n=1}^{N} \sum_{t=1}^{T} \left( \mathcal{L}_{\text{geom}} + \mathcal{L}_{\text{app}} \right).
\end{equation}

\section{EXPERIMENTS}
 
\subsection{Experimental Setup}
 
\textbf{Dataset.} CARLA-NVS is split into 15 training scenes and 5 held-out test scenes covering urban, rural, and highway environments. GS-Net is trained only on the training scenes, and all rendering results are reported on the held-out scenes under the protocols in Sec.~\ref{sec:eval_protocol}. \textcolor{black}{To better assess the sim-to-real domain gap highlighted by recent simulation and digital-twin driving datasets~\cite{driving,truecity}, we additionally evaluate on real nuScenes sequences~\cite{caesar2020nuscenes} using its six surround-view cameras and a CARLA-NVS-aligned split. Since nuScenes has a fixed sensor layout and no ground truth at extrapolated viewpoints, we report same-sensor quantitative results and cross-sensor qualitative results.} Metrics are Peak Signal-to-Noise Ratio (PSNR), Structural Similarity Index Measure (SSIM)~\cite{wang2004ssim}, and Learned Perceptual Image Patch Similarity (LPIPS)~\cite{zhang2018lpips}.
 
\textcolor{black}{\textbf{Baselines.} We compare GS-Net against the standard 3DGS~\cite{c2}, the densification-enhanced variant GaussianPro~\cite{cheng2024Gaussianpro}, the NeRF-style method Nerfacto~\cite{nerfacto}, and two feed-forward generalizable methods, MVSplat~\cite{mvsplat} and MonoSplat~\cite{monosplat}, the last two evaluated both zero-shot and after fine-tuning on CARLA-NVS. All baselines use the same data splits and SfM poses as GS-Net.}

\textbf{Implementation.} GS-Net is trained for 200 epochs with a batch size of 512, optimized via Adam with a learning rate of $1 \times 10^{-3}$. The expansion factor is set to $T=5$ and the neighborhood size to $M=3$. The scene-level optimization of 3DGS~\cite{c2} follows its default configuration. Both training and inference are conducted on a single Tesla T4 GPU.

\subsection{Main Results}
 
\begin{figure}[t]
    \centering
    \savebox{\panelbox}{\includegraphics[width=\dimexpr\linewidth-1.3em-2pt\relax]{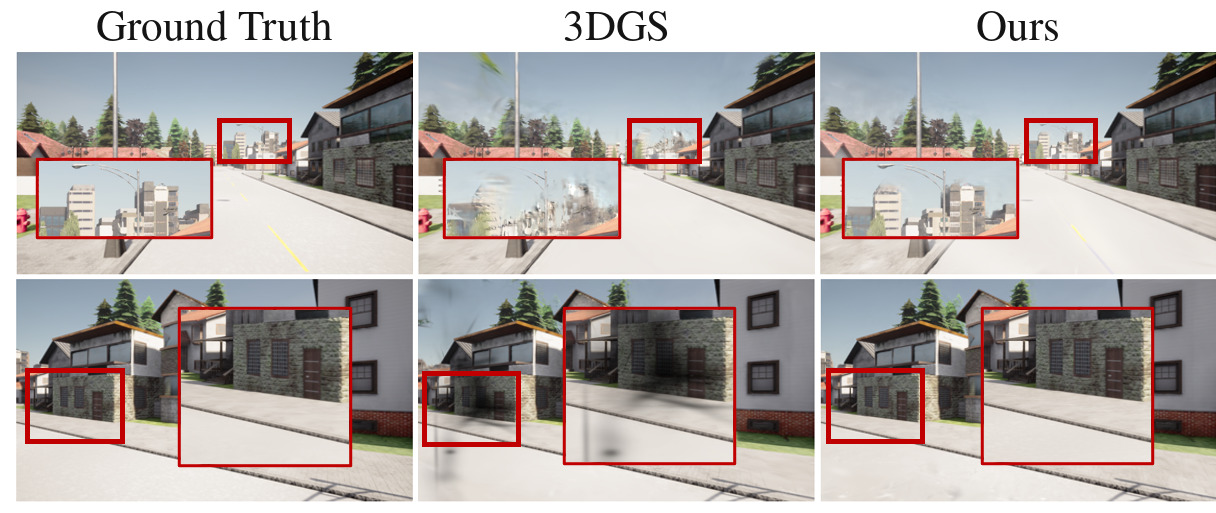}}%
    \makebox[1.3em][r]{\raisebox{0.46\ht\panelbox}{\raisebox{-0.5\height}{\small(a)}}}\hspace{2pt}\usebox{\panelbox}\\[6pt]
    \savebox{\panelbox}{\includegraphics[width=\dimexpr\linewidth-1.3em-2pt\relax]{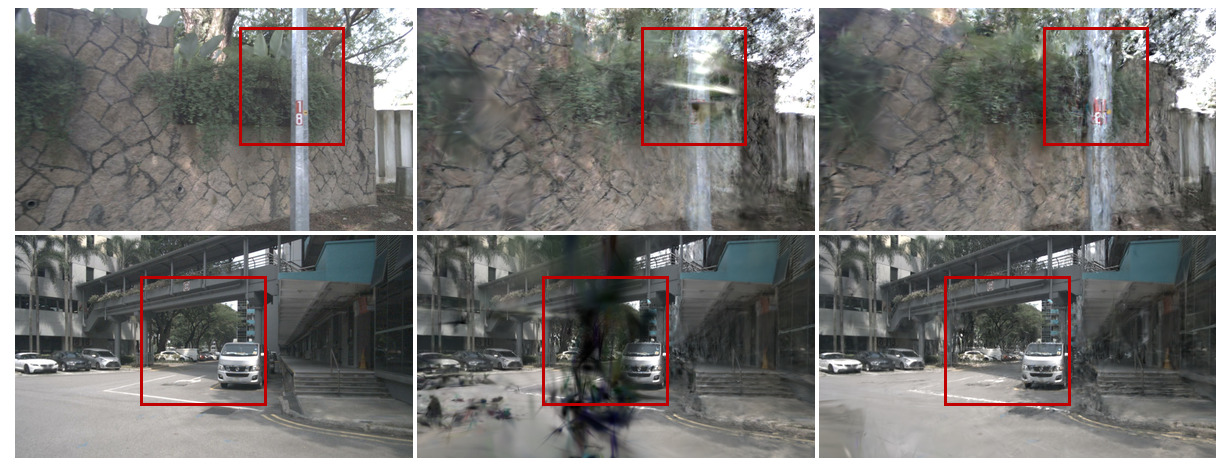}}%
    \makebox[1.3em][r]{\raisebox{0.5\ht\panelbox}{\raisebox{-0.5\height}{\small(b)}}}\hspace{2pt}\usebox{\panelbox}
   \caption{Qualitative comparison under the Same-Sensor Evaluation (SSE) protocol. Red boxes mark the highlighted regions. (a) On CARLA-NVS, compared with 3DGS, GS-Net recovers sharper details and fewer artifacts in textureless and distant regions. (b) \textcolor{black}{On nuScenes sequences, the 3DGS baseline produces hazy floating artifacts that blur the underlying structure, whereas GS-Net suppresses these artifacts and recovers cleaner geometry on real reconstructions.}}
    \label{fig:sse_qual}
\end{figure}

\textbf{Same-Sensor Evaluation.} We first evaluate GS-Net under the SSE protocol, with results reported in Table~\ref{tab:sse_results}. SfM reconstructions tend to be sparse in textureless regions such as building facades and uniform pavement, leaving the standard 3DGS baseline with insufficient geometric coverage. GS-Net densifies the sparse initialization, raising the average PSNR by 2.08\,dB with consistent improvements in SSIM and LPIPS across all five test scenes. To further verify its plug-and-play compatibility, we integrate GS-Net with GaussianPro~\cite{cheng2024Gaussianpro}. GaussianPro refines the optimization process of 3DGS but still relies on the same sparse SfM initialization, yielding an average PSNR of 26.40\,dB. Replacing its initialization with GS-Net brings the performance to 28.52\,dB, a gain of 2.12\,dB. This result demonstrates that GS-Net can be integrated with advanced 3DGS variants as a plug-and-play initialization module. \textcolor{black}{To verify that the improvement is not specific to simulation, we further evaluate GS-Net on real nuScenes sequences under the same-sensor protocol. This experiment uses camera poses and sparse SfM point clouds reconstructed from real images, and therefore tests whether GS-Net can be integrated into a real reconstruction pipeline. As reported in the nuScenes block of Table~\ref{tab:sse_results}, GS-Net improves the average PSNR from 20.08 to 20.37\,dB over the 3DGS baseline, and the qualitative comparison in Fig.~\ref{fig:sse_qual}(b) shows that GS-Net suppresses the hazy floating artifacts of 3DGS in the highlighted regions. These results support that the learned Gaussian initialization prior remains effective on real-world reconstructions.}

\textbf{Cross-Sensor Evaluation.} We further evaluate cross-sensor generalization under the CSE protocol, where test views come from camera positions entirely absent during reconstruction. As reported in Table~\ref{tab:cse_results}, the 3DGS baseline drops from 26.22 to 20.98\,dB relative to the same-sensor protocol, reflecting the increased difficulty of spatial extrapolation over temporal interpolation. \textcolor{black}{As a NeRF-style reference, Nerfacto~\cite{nerfacto} degrades to 20.34\,dB under CSE by a margin comparable to 3DGS, indicating that the difficulty of the cross-sensor regime arises from the extrapolation itself rather than from a property of the 3DGS pipeline. Feed-forward generalizable methods achieve lower scores under this cross-sensor extrapolation protocol. MVSplat~\cite{mvsplat} and MonoSplat~\cite{monosplat} fall below 3DGS without adaptation, and after fine-tuning only MonoSplat slightly exceeds the 3DGS baseline in PSNR while staying worse in SSIM and LPIPS. Inferred once from the source views, their predicted geometry degrades where the target viewpoint leaves the observed coverage. GS-Net instead acts at initialization, expanding the sparse points into a denser, more complete set of Gaussians that gives the per-scene optimization a reliable starting geometry in these regions. GS-Net+3DGS achieves the best scores on all three metrics, outperforming 3DGS by 1.86\,dB and the strongest fine-tuned baseline by 1.51\,dB in PSNR. These results indicate that GS-Net provides a transferable geometric prior rather than overfitting to the source camera trajectories.}

\begin{figure*}[t]
\centering
\includegraphics[width=\linewidth]{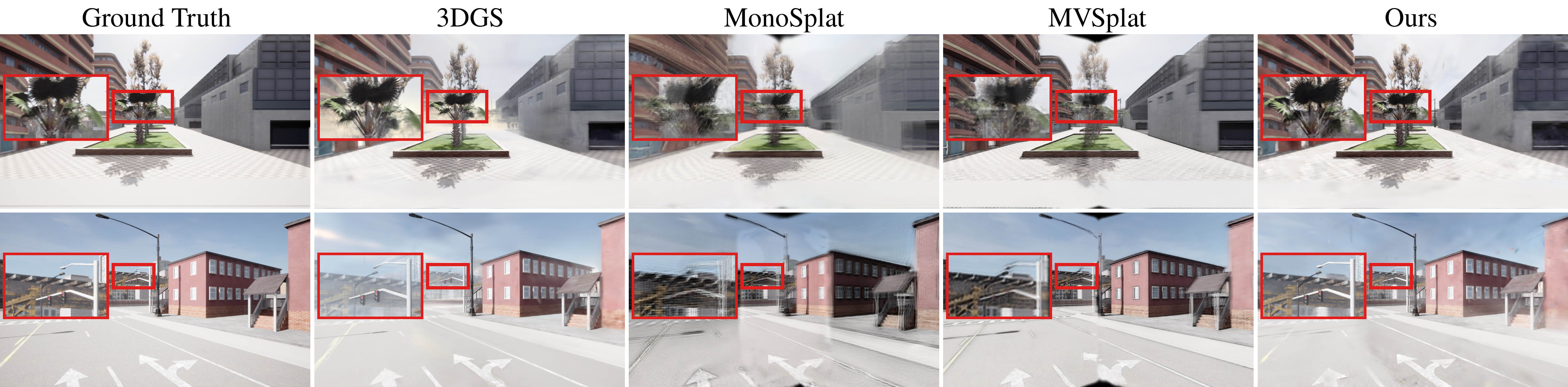}
\caption{\textcolor{black}{Qualitative comparison under the Cross-Sensor Evaluation (CSE) protocol on CARLA-NVS. Red boxes mark the compared regions. At the unseen target viewpoints the feed-forward baselines produce haze, blur, and distorted geometry, whereas GS-Net stays closest to the ground truth.}}
\label{fig:cse_qual}
\end{figure*}

\begin{figure*}[t]
\centering
\includegraphics[width=\linewidth]{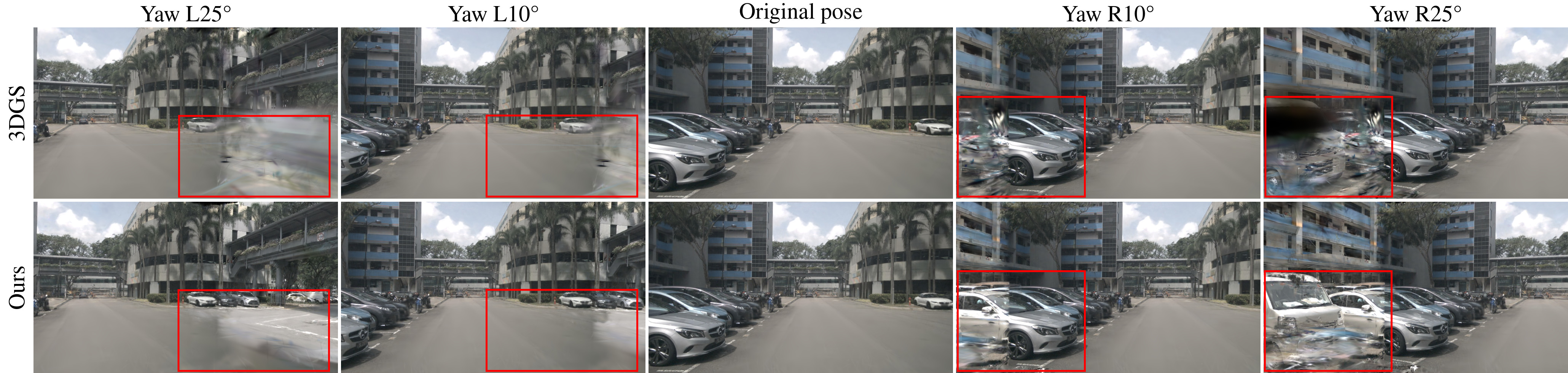}
\caption{\textcolor{black}{Qualitative cross-sensor comparison on a real nuScenes urban street scene. The top row shows 3DGS and the bottom row shows GS-Net, rendered from the original captured pose and from poses yawed left and right by $10^{\circ}$ and $25^{\circ}$. The original captured pose serves as an observed-view reference for the rotation sequence. Red boxes mark newly exposed regions at rotated viewpoints, where 3DGS produces ghosting and floating artifacts while GS-Net maintains more coherent structure.}}
\label{fig:nusc_real}
\end{figure*}

\begin{table}[t]
\centering
\caption{Quantitative Results Under the Same-Sensor Evaluation (SSE) Protocol \textcolor{black}{on CARLA-NVS and nuScenes}, Best Results Highlighted in \textbf{Bold}.}
\label{tab:sse_results}
\resizebox{\linewidth}{!}{
\begin{tabular}{@{}ll ccccc c@{}}
\toprule
\textbf{Metric} & \textbf{Method} & \textbf{S1} & \textbf{S2} & \textbf{S3} & \textbf{S4} & \textbf{S5} & \textbf{Avg.} \\
\midrule
\multicolumn{8}{@{}l}{\textcolor{black}{\textit{CARLA-NVS}}} \\
\addlinespace[2pt]
\multirow{5}{*}{PSNR $\uparrow$} 
& \textcolor{black}{Nerfacto} & \textcolor{black}{24.58} & \textcolor{black}{25.55} & \textcolor{black}{25.96} & \textcolor{black}{22.37} & \textcolor{black}{24.15} & \textcolor{black}{24.52} \\
& 3DGS & 26.80 & 27.36 & 27.95 & 24.35 & 24.66 & 26.22 \\
& \textbf{GS-Net+3DGS} & \textbf{29.62} & \textbf{28.43} & \textbf{30.06} & \textbf{28.08} & \textbf{25.32} & \textbf{28.30} \\
\cmidrule{2-8}
& GaussianPro & 26.91 & 27.52 & 28.21 & 24.61 & 24.76 & 26.40 \\
& \textbf{GS-Net+GaussianPro} & \textbf{29.27} & \textbf{29.19} & \textbf{30.56} & \textbf{28.35} & \textbf{25.22} & \textbf{28.52} \\
\midrule
\multirow{5}{*}{SSIM $\uparrow$} 
& \textcolor{black}{Nerfacto} & \textcolor{black}{0.865} & \textcolor{black}{0.911} & \textcolor{black}{0.885} & \textcolor{black}{0.841} & \textcolor{black}{0.815} & \textcolor{black}{0.863} \\
& 3DGS & 0.929 & 0.942 & 0.941 & 0.915 & 0.867 & 0.919 \\
& \textbf{GS-Net+3DGS} & \textbf{0.947} & \textbf{0.946} & \textbf{0.954} & \textbf{0.936} & \textbf{0.870} & \textbf{0.931} \\
\cmidrule{2-8}
& GaussianPro & 0.933 & 0.936 & 0.944 & 0.927 & 0.868 & 0.922 \\
& \textbf{GS-Net+GaussianPro} & \textbf{0.948} & \textbf{0.943} & \textbf{0.953} & \textbf{0.938} & \textbf{0.873} & \textbf{0.931} \\
\midrule
\multirow{5}{*}{LPIPS $\downarrow$} 
& \textcolor{black}{Nerfacto} & \textcolor{black}{0.190} & \textcolor{black}{0.156} & \textcolor{black}{0.176} & \textcolor{black}{0.207} & \textcolor{black}{0.228} & \textcolor{black}{0.192} \\
& 3DGS & 0.166 & 0.152 & 0.152 & 0.225 & 0.202 & 0.179 \\
& \textbf{GS-Net+3DGS} & \textbf{0.131} & \textbf{0.139} & \textbf{0.141} & \textbf{0.160} & \textbf{0.192} & \textbf{0.153} \\
\cmidrule{2-8}
& GaussianPro & 0.147 & 0.146 & 0.139 & 0.179 & 0.199 & 0.162 \\
& \textbf{GS-Net+GaussianPro} & \textbf{0.129} & \textbf{0.142} & \textbf{0.121} & \textbf{0.157} & \textbf{0.191} & \textbf{0.148} \\
\midrule
\multicolumn{8}{@{}l}{\textcolor{black}{\textit{nuScenes}}} \\
\addlinespace[2pt]
\multirow{2}{*}{\textcolor{black}{PSNR $\uparrow$}}
& \textcolor{black}{3DGS} & \textcolor{black}{19.93} & \textcolor{black}{22.04} & \textcolor{black}{19.46} & \textcolor{black}{\textbf{18.86}} & \textcolor{black}{20.09} & \textcolor{black}{20.08} \\
& \textcolor{black}{\textbf{GS-Net+3DGS}} & \textcolor{black}{\textbf{20.28}} & \textcolor{black}{\textbf{22.35}} & \textcolor{black}{\textbf{19.89}} & \textcolor{black}{18.80} & \textcolor{black}{\textbf{20.53}} & \textcolor{black}{\textbf{20.37}} \\
\cmidrule{2-8}
\multirow{2}{*}{\textcolor{black}{SSIM $\uparrow$}}
& \textcolor{black}{3DGS} & \textcolor{black}{0.807} & \textcolor{black}{0.765} & \textcolor{black}{0.651} & \textcolor{black}{\textbf{0.710}} & \textcolor{black}{0.755} & \textcolor{black}{0.738} \\
& \textcolor{black}{\textbf{GS-Net+3DGS}} & \textcolor{black}{\textbf{0.808}} & \textcolor{black}{\textbf{0.767}} & \textcolor{black}{\textbf{0.659}} & \textcolor{black}{0.704} & \textcolor{black}{\textbf{0.760}} & \textcolor{black}{\textbf{0.740}} \\
\cmidrule{2-8}
\multirow{2}{*}{\textcolor{black}{LPIPS $\downarrow$}}
& \textcolor{black}{3DGS} & \textcolor{black}{0.315} & \textcolor{black}{0.328} & \textcolor{black}{0.398} & \textcolor{black}{\textbf{0.416}} & \textcolor{black}{0.363} & \textcolor{black}{\textbf{0.364}} \\
& \textcolor{black}{\textbf{GS-Net+3DGS}} & \textcolor{black}{\textbf{0.309}} & \textcolor{black}{\textbf{0.327}} & \textcolor{black}{\textbf{0.394}} & \textcolor{black}{0.433} & \textcolor{black}{\textbf{0.360}} & \textcolor{black}{0.365} \\
\bottomrule
\end{tabular}
}
\end{table}

\begin{table}[t]
\centering
\caption{\textcolor{black}{Quantitative Results Under the Cross-Sensor Evaluation (CSE) Protocol on CARLA-NVS, Averaged Over the Five Test Scenes. Best Results in \textbf{Bold}, Second Best \underline{Underlined}.}}
\label{tab:cse_results}
\resizebox{\linewidth}{!}{
\begin{tabular}{@{}llccc@{}}
\toprule
\textbf{Method} & \textbf{Paradigm} & \textbf{PSNR}\,$\uparrow$ & \textbf{SSIM}\,$\uparrow$ & \textbf{LPIPS}\,$\downarrow$ \\
\midrule
\textcolor{black}{Nerfacto~\cite{nerfacto}} & \textcolor{black}{NeRF} & \textcolor{black}{20.34} & \textcolor{black}{0.729} & \textcolor{black}{0.277} \\
3DGS~\cite{c2} & per-scene 3DGS & 20.98 & \underline{0.852} & \underline{0.272} \\
\midrule
\textcolor{black}{MVSplat (zero-shot)~\cite{mvsplat}} & \textcolor{black}{feed-forward} & \textcolor{black}{20.20} & \textcolor{black}{0.784} & \textcolor{black}{0.382} \\
\textcolor{black}{MonoSplat (zero-shot)~\cite{monosplat}} & \textcolor{black}{feed-forward} & \textcolor{black}{19.62} & \textcolor{black}{0.807} & \textcolor{black}{0.413} \\
\textcolor{black}{MVSplat (fine-tuned)~\cite{mvsplat}} & \textcolor{black}{feed-forward} & \textcolor{black}{20.88} & \textcolor{black}{0.787} & \textcolor{black}{0.341} \\
\textcolor{black}{MonoSplat (fine-tuned)~\cite{monosplat}} & \textcolor{black}{feed-forward} & \textcolor{black}{\underline{21.33}} & \textcolor{black}{0.816} & \textcolor{black}{0.358} \\
\midrule
\textbf{GS-Net+3DGS (Ours)} & hybrid & \textbf{22.84} & \textbf{0.855} & \textbf{0.249} \\
\bottomrule
\end{tabular}
}
\end{table}

\subsection{Qualitative Results}
\label{sec:qualitative}

\textcolor{black}{\textbf{CARLA-NVS.}} Fig.~\ref{fig:sse_qual}(a) shows that GS-Net reduces blurry textures and floating artifacts in distant or textureless regions under SSE. \textcolor{black}{Fig.~\ref{fig:cse_qual} further shows the CSE setting, where feed-forward baselines produce haze, blur, and distorted geometry at unseen target viewpoints, while GS-Net remains closest to the ground truth.}

\textcolor{black}{\textbf{nuScenes.} We further examine the cross-sensor behavior on real data. As nuScenes offers no real reference beyond its fixed layout, we render each scene from the original captured pose and from poses yawed left and right by $10^{\circ}$ and $25^{\circ}$, and compare 3DGS with GS-Net. The original captured pose serves as an observed-view reference for the rotation sequence. As shown in Fig.~\ref{fig:nusc_real}, as the yaw angle grows, the regions newly exposed by the rotation fall outside the original coverage, where 3DGS produces ghosting and floating artifacts that worsen from $10^{\circ}$ to $25^{\circ}$, as marked by the red boxes. GS-Net suppresses these artifacts in both rotation directions and maintains structural coherence.}

\subsection{\textcolor{black}{Heterogeneous Vehicle Data Reuse Validation}}
\begin{figure}[t]
\centering
\includegraphics[width=\linewidth]{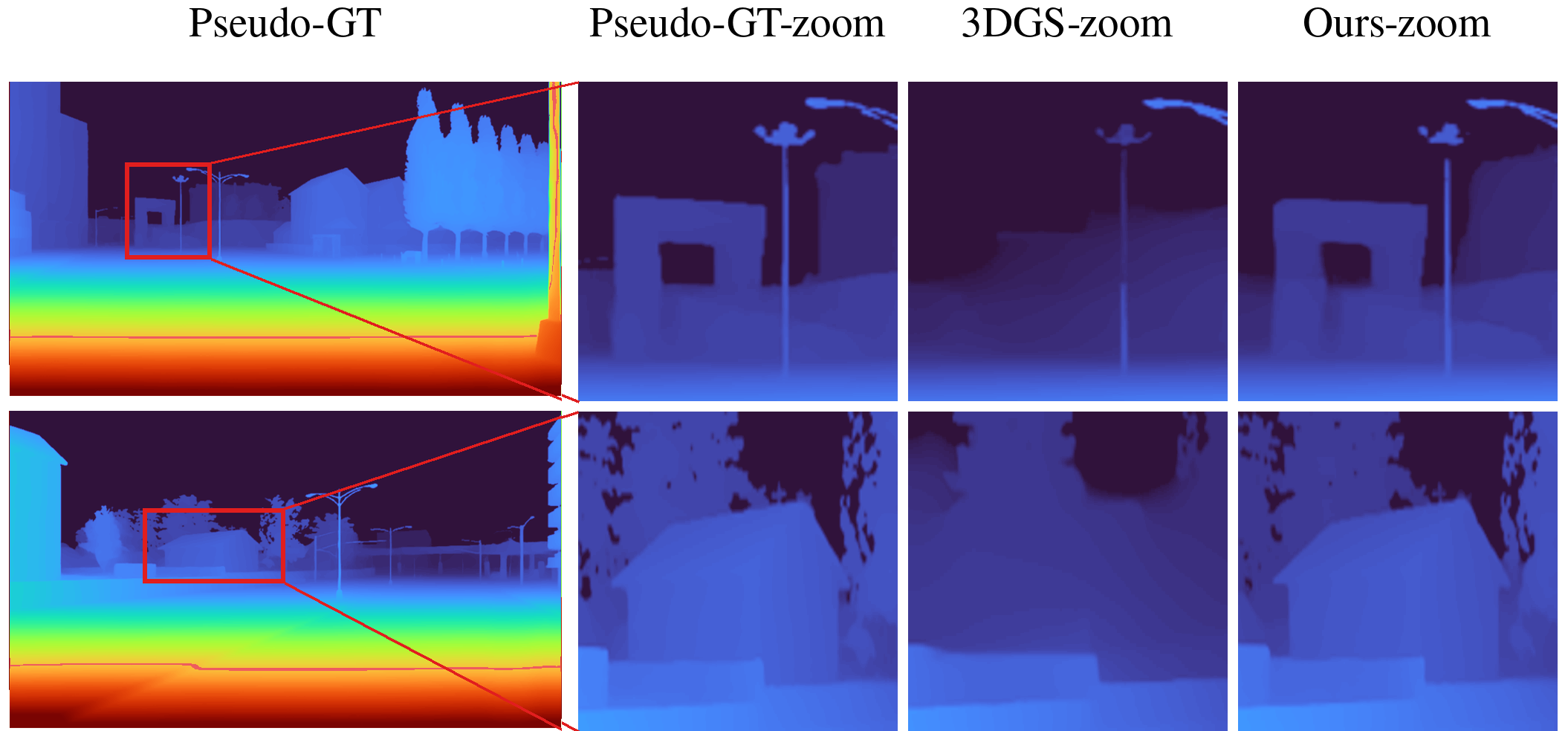}
\caption{\textcolor{black}{Downstream depth consistency under the CSE protocol on CARLA-NVS. For each example we show the Pseudo-GT depth map with a red box, then the magnified box for the Pseudo-GT, the 3DGS baseline, and GS-Net. The Pseudo-GT is a frozen monocular-depth model's prediction on the captured target image, and a result closer to it is better. At extrapolated target views, 3DGS produces distorted depth, whereas GS-Net stays close to the Pseudo-GT.}}
\label{fig:downstream_qual}
\end{figure}

\textcolor{black}{Cross-sensor data reuse aims to transform data captured under one sensor layout into views usable under another layout. Beyond visual fidelity, such views should preserve the information required by downstream perception models. Under the CSE protocol, we therefore test whether a target-layout view synthesized from source-layout data induces perception outputs consistent with those from the real target capture. For each held-out target view, we render the scene from the source-layout reconstruction using both 3DGS and GS-Net, and run a frozen monocular-depth model, Depth-Anything-V2~\cite{depthanything}, on each rendering and on the real capture. We do not score these predictions against CARLA ground truth directly, because the depth model predicts relative, scale-ambiguous depth that is not aligned with CARLA's absolute depth. We therefore use the predictions on the real capture as Pseudo-GT and measure how closely the rendering predictions match them, where higher agreement indicates a more reusable view. As shown in Table~\ref{tab:downstream} and Fig.~\ref{fig:downstream_qual}, GS-Net matches the Pseudo-GT more closely than 3DGS on both depth metrics. GS-Net therefore better preserves the geometric cues required by downstream perception, making the synthesized cross-sensor views more functionally reusable.}

\begin{table}[t]
\centering
\caption{\textcolor{black}{Downstream depth consistency under CSE on CARLA-NVS, measured as the agreement between a frozen depth model's predictions on the rendered view and on the captured target (Pseudo-GT). Best in \textbf{Bold}. Green shows GS-Net's improvement over 3DGS.}}
\label{tab:downstream}
\begin{tabular*}{\linewidth}{@{\extracolsep{\fill}}lcc}
\toprule
\textbf{Method} & AbsRel\,$\downarrow$ & $\delta<1.25$\,$\uparrow$\\
\midrule
3DGS & 0.1104 & 93.04\\
\textbf{GS-Net+3DGS (Ours)} & \textbf{0.0843}\,{\scriptsize\boldmath\bfseries\textcolor{green!60!black}{($\downarrow$\kern1pt 23.6\%)}} & \textbf{94.70}\,{\scriptsize\boldmath\bfseries\textcolor{green!60!black}{($\uparrow$\kern-3.5pt 1.66\,pp)}}\\
\bottomrule
\end{tabular*}
\end{table}

\subsection{Ablation Studies}

\textbf{Efficiency Analysis.} Beyond rendering quality, we compare the computational overhead of enhancing the sparse SfM initialization in Table~\ref{tab:efficiency}. Replacing SfM points with dense MVS reconstructions improves PSNR to 28.50\,dB but introduces 102 minutes of iterative dense matching and fusion. GS-Net achieves a comparable PSNR of 28.30\,dB with an enhancement time of only two minutes, representing a $50\times$ speedup over MVS in the enhancement stage. This efficiency stems from the feed-forward design of GS-Net, which amortizes the cost of geometric densification across training scenes and requires only a single forward pass at inference.

\begin{table*}[t]
\centering
\begin{minipage}[t]{0.48\linewidth}
\centering
\caption{Efficiency Comparison Excluding Shared SfM Time. PC Densify and Ellip. Infer. denote point cloud densification and Gaussian ellipsoid inference, respectively. The best results are in \textbf{Bold} and the second best are \underline{Underlined}}
\label{tab:efficiency}
\resizebox{\linewidth}{!}{%
\begin{tabular}{@{} l l c c c c @{}}
\toprule
\multirow{2}{*}{\textbf{Method}} & \multirow{2}{*}{\textbf{Type}} & \textbf{Enhance} & \textbf{Optim.} & \textbf{Total} & \multirow{2}{*}{\textbf{PSNR $\uparrow$}} \\
 & & \textbf{(min)} & \textbf{(min)} & \textbf{(min)} & \\
\midrule
SfM (Baseline) & -- & -- & 45 & \underline{45} & 26.22 \\
\quad + MVS & PC Densify & 102 & \textbf{30} & 132 & \textbf{28.50} \\
\quad + \textbf{GS-Net (Ours)} & Ellip. Infer. & \textbf{2} & \underline{42} & \textbf{44} & \underline{28.30} \\
\bottomrule
\end{tabular}%
}
\end{minipage}\hfill
\begin{minipage}[t]{0.48\linewidth}
\centering
\caption{Ablation Study on the Predicted Gaussian Attributes, Best Results Highlighted in \textbf{Bold}}
\label{tab:ablation_params}
\resizebox{\linewidth}{!}{%
\begin{tabular}{@{}l cc@{}}
\toprule
\textbf{Configuration} & \textbf{SSE (PSNR)} & \textbf{CSE (PSNR)} \\
\midrule
3DGS baseline & 26.22 & 20.98 \\
+ $\Delta\mu$ & 27.18 {\scriptsize (+0.96)} & 21.76 {\scriptsize (+0.78)} \\
+ $\Delta\mu$ + $\Delta C_{\text{rgb}}$ & 27.71 {\scriptsize (+1.49)} & 22.19 {\scriptsize (+1.21)} \\
+ $\Delta\mu$ + $\Delta C_{\text{rgb}}$ + $\alpha$ & 28.02 {\scriptsize (+1.80)} & 22.51 {\scriptsize (+1.53)} \\
+ $\Delta\mu$ + $\Delta C_{\text{rgb}}$ + $\alpha$ + $S, R$ (Full) & \textbf{28.30} {\scriptsize \textbf{(+2.08)}} & \textbf{22.84} {\scriptsize \textbf{(+1.86)}} \\
\bottomrule
\end{tabular}%
}
\end{minipage}
\end{table*}

\textbf{Contribution of Individual Attributes.} We isolate each predicted attribute by progressively adding it to the GS-Net output (Table~\ref{tab:ablation_params}). The spatial offset $\Delta\mu$ gives the largest single gain, +0.96\,dB on SSE and +0.78\,dB on CSE, confirming that accurate primitive positioning is the most critical factor. The color offset $\Delta C_{\text{rgb}}$ and opacity $\alpha$ add further improvements by easing per-scene photometric optimization, and the full parameterization with scale $S$ and rotation $R$ performs best.

\section{CONCLUSION}
This paper presents GS-Net, a lightweight plug-and-play module for cross-sensor view synthesis in autonomous driving. GS-Net learns a cross-scene initialization prior that expands sparse SfM points into dense Gaussian primitives, improving 3DGS rendering under both same-sensor interpolation and cross-sensor extrapolation. To quantify this setting, we introduce CARLA-NVS, a controlled benchmark that provides ground-truth images at alternative sensor placements unavailable in fixed-layout real driving datasets. We further extend the evaluation to real nuScenes sequences and downstream perception consistency, showing that GS-Net remains effective on real-world reconstructions and preserves cues useful for cross-vehicle data reuse.


\end{document}